\documentclass[%
 preprint,
 amsmath,amssymb,
]{revtex4-1}

\usepackage{graphicx}% Include figure files
\usepackage{dcolumn}% Align table columns on decimal point
\usepackage{bm}% bold math
\usepackage{hyperref}% add hypertext capabilities
\usepackage{amsmath}
\usepackage{multirow}
\usepackage[utf8]{inputenc}
\usepackage{booktabs}
\usepackage{setspace}

\usepackage{color}
\definecolor{redcolor}{rgb}{1.0,0.,0.}

\begin{document}

\preprint{}

%\title{Using word embeddings to improve the discriminability of co-occurrence text networks%: application %to authorship attribution
%}% Force line breaks with \\

\title{Using word embeddings to improve the discriminability of co-occurrence text networks}

\author{Laura V. C. Quispe}
\author{Jorge A. V. Tohalino}%
\author{Diego R. Amancio}%
\email{diego@icmc.usp.br}
\affiliation{%
Institute of Mathematics and Computer Science, Department of Computer Science, University of S\~{a}o Paulo,
S\~{a}o Carlos, SP,  Brazil
}%


\date{\today}% It is always \today, today,

\begin{abstract}
Word co-occurrence networks have been employed to analyze texts both in the practical and theoretical scenarios. Despite the relative success in several applications, traditional co-occurrence networks fail in establishing links between similar words whenever they appear distant in the text. Here we investigate whether the use of word embeddings as a tool to create virtual links in co-occurrence networks may improve the quality of classification systems. Our results revealed that the discriminability in the stylometry task is improved when using \emph{Glove}, \emph{Word2Vec} and \emph{FastText}. In addition, we found that optimized results are obtained when \emph{stopwords} are not disregarded and a simple global thresholding strategy is used to establish virtual links. Because the proposed approach is able to improve the representation of texts as complex networks, we believe that it could be extended to study other natural language processing tasks. Likewise, theoretical languages studies could benefit from the adopted enriched representation of word co-occurrence networks.
\end{abstract}

\maketitle

%\tableofcontents

\section{Introduction}

The ability to construct complex and diverse linguistic structures is one of the main features that set us apart from all other species. Despite its ubiquity, some language aspects remain unknown. Topics such as language origin and evolution
have been studied by researchers from diverse disciplines, including Linguistic, Computer Science, Physics and Mathematics~\cite{faggian2019synchronization,kong2019skill,shimada2019formation}. In order to better understand the underlying language mechanisms and universal linguistic properties, several models have been developed~\cite{miller1994hidden,baronchelli2013networks}. A particular language representation regards texts as complex systems~\cite{cong2014approaching}. Written texts can be considered as complex networks (or graphs), where nodes could represent syllables, words, sentences, paragraphs or even larger chunks~\cite{cong2014approaching}. In such models, network edges represent the proximity between nodes, e.g. the frequency of the co-occurrence of words. Several interesting results have been obtained from networked models, such as the explanation of Zipf's Law as a consequence of the least effort principle and theories on the nature of syntactical relationships~\cite{i2003least,i2006syntactic}.

In a more practical scenario, text networks have been used in text classification tasks~\cite{amancio2015complex,MEHRI20122429,segarra2015authorship}. The main advantage of the model is that it does not rely on deep semantical information to obtain competitive results. Another advantage of graph-based approaches is that, when combined with other approaches, it yields competitive  results~\cite{santos2017enriching}. A simple, yet recurrent text model is the well-known word co-occurrence network. After optional textual pre-processing steps, in a co-occurrence network each different word becomes a node and edges are established via co-occurrence in a desired window. A common strategy connects only adjacent words in the so called word adjacency networks.

While the co-occurrence representation yields good results in classification scenarios, some important features are not considered in the model. For example, long-range syntactical links, though less frequent than adjacent syntactical relationships, might be disregarded from a simple word adjacency approach~\cite{i2004patterns}. In addition, semantically similar words not sharing the same lemma are mapped into distinct nodes. In order to address these issues, here we introduce a modification of the traditional network representation by establishing additional edges, referred to as ``\emph{virtual}'' edges. In the proposed model, in addition to the co-occurrence edges, we link two nodes (words) if the corresponding word embedding representation is similar. While this approach still does not merge similar nodes into the same concept, similar nodes are explicitly linked via virtual edges.

Our main objective here is to evaluate whether such an approach is able to improve the discriminability of word co-occurrence networks in a typical text network classification task. We evaluate the methodology for different embedding techniques, including \emph{GloVe}, \emph{Word2Vec} and \emph{FastText}. We also investigated different thresholding strategies to establish virtual links. Our results revealed, as a proof of principle, that the proposed approach is able to improve the discriminability of the classification when compared to the traditional co-occurrence network. While the gain in performance depended upon the text length being considered, we found relevant gains for intermediary text lengths. Additional results also revealed that a simple thresholding strategy combined with the use of stopwords tends to yield the best results.

We believe that the proposed representation could be applied in other text classification tasks, which could lead to potential gains in performance. Because the inclusion of virtual edges is a simple technique to make the network denser, such an approach can benefit networked representations with a limited number of nodes and edges.
This representation could also shed light into language mechanisms in theoretical studies relying on the representation of text as complex networks. Potential novel research lines leveraging the adopted approach to improve the characterization of texts in other applications are presented in the conclusion.

%This introduces two main advantages,
%- %Falar que nao comparou com o estado porque o objetivo eh fornecer um metodo para aperfeicoar.

\section{Related works}

Complex networks have been used in a wide range of fields, including in Social Sciences~\cite{borgatti2009network}, Neuroscience~\cite{van2010comparing}, Biology~\cite{rodrigues2011resilience}, Scientometry~\cite{zeng2017science} and Pattern Recognition~\cite{breve2013fuzzy,breve2019interactive,breve2017building,barbieri2011entropy}. In text analysis, networks are used to uncover language patterns, including the origins of the ever present Zipf's Law~\cite{ferrer2003least} and the analysis of linguistic properties of natural and unknown texts~\cite{estevez2019complexity,montemurro2013keywords}. Applications of network science in text mining and text classification encompasses applications in semantic analysis~\cite{hassan2007random,correa2018word,10.1371/journal.pone.0222870,stella2015patterns}, authorship attribution~\cite{STANISZ2019301,chen2018does} and stylometry~\cite{STANISZ2019301,GAO2014579,garg2018structure}. Here we focus in the stylometric analysis of texts using complex networks.

In~\cite{STANISZ2019301}, the authors used a co-occurrence network to study a corpus of English and Polish books. They considered a dataset of 48 novels, which were written by 8 different authors. Differently from traditional co-occurrence networks, some punctuation marks were considered as words when mapping texts as networks. The authors also decided to create a methodology to normalize the obtained network metrics, since they considered documents with variations in length. A similar approach was adopted in a similar study~\cite{2015concentric}, with a focus on comparing novel measurements and measuring the effect of considering stopwords in the network structure.

A different approach to analyze co-occurrence networks was devised in~\cite{marinho2017labelled}. Whilst most approaches only considered traditional network measurements or devised novel topological and dynamical measurements, the authors combined networked and semantic information to improve the performance of network-based classification. Interesting, the combined use of network motifs and node labels (representing the corresponding words) allowed an improvement in performance in the considered task. A similar combination of techniques using a hybrid approach was proposed in~\cite{amancio2015complex}. Networked-based approaches has also been applied to the authorship recognition tasks in other languages, including Persian texts~\cite{MEHRI20122429}.

Co-occurrence networks have been used in other contexts other than stylometric analysis. The main advantage of this approach is illustrated in the task aimed at diagnosing diseases via text analysis~\cite{santos2017enriching}. Because the topological analysis of co-occurrence language networks do not require deep semantic analysis, this model is able to model text created by patients suffering from cognitive impairment~\cite{santos2017enriching}. Recently, it has been shown that the combination of network and traditional features could be used to improve the diagnosis of patients with cognitive impairment~\cite{santos2017enriching}. Interestingly, this was one of the first approaches suggesting the use of embeddings to address the particular problem of lack of statistics to create a co-occurrence network in short documents~\cite{amancio2015probing}.

While many of the works dealing with word co-occurrence networks have been proposed in the last few years, no systematic study of the effects of including information from word embeddings in such networks has been analyzed. This work studies how links created via embeddings information modify the underlying structure of networks and, most importantly, how it can improve the model to provide improved classification performance in the stylometry task.

\section{Material and Methods}

To represent texts as networks, we used the so-called word adjacency network representation~\cite{akimushkin2017text,STANISZ2019301,2015concentric}. Typically, before creating the networks, the text is pre-processed. An optional pre-processing step is the removal of \emph{stopwords}. This step is optional because such words include mostly article and prepositions, which may be artlessly represented by network edges. However, in some applications -- including the authorship attribution task -- stopwords (or \emph{function words}) play an important role in the stylistic characterization of texts~\cite{2015concentric}. A list of stopwords considered in this study is available in the Supplementary Information.
%\textcolor{red}{Table ?? of the Supplementary Information}.

The pre-processing step may also include a lemmatization procedure. This step aims at mapping words conveying the same meaning into the same node. In the lemmatization process, nouns and verbs are mapped into their singular and infinite forms. Note that, while this step is useful to merge words sharing a \emph{lemma} into the same node, more complex semantical relationships are overlooked. For example, if ``car'' and ``vehicle'' co-occur in the same text, they are considered as distinct nodes, which may result in an inaccurate representation of the text.

\begin{figure}%[h]
    \centering
    \includegraphics[width=0.8\textwidth]{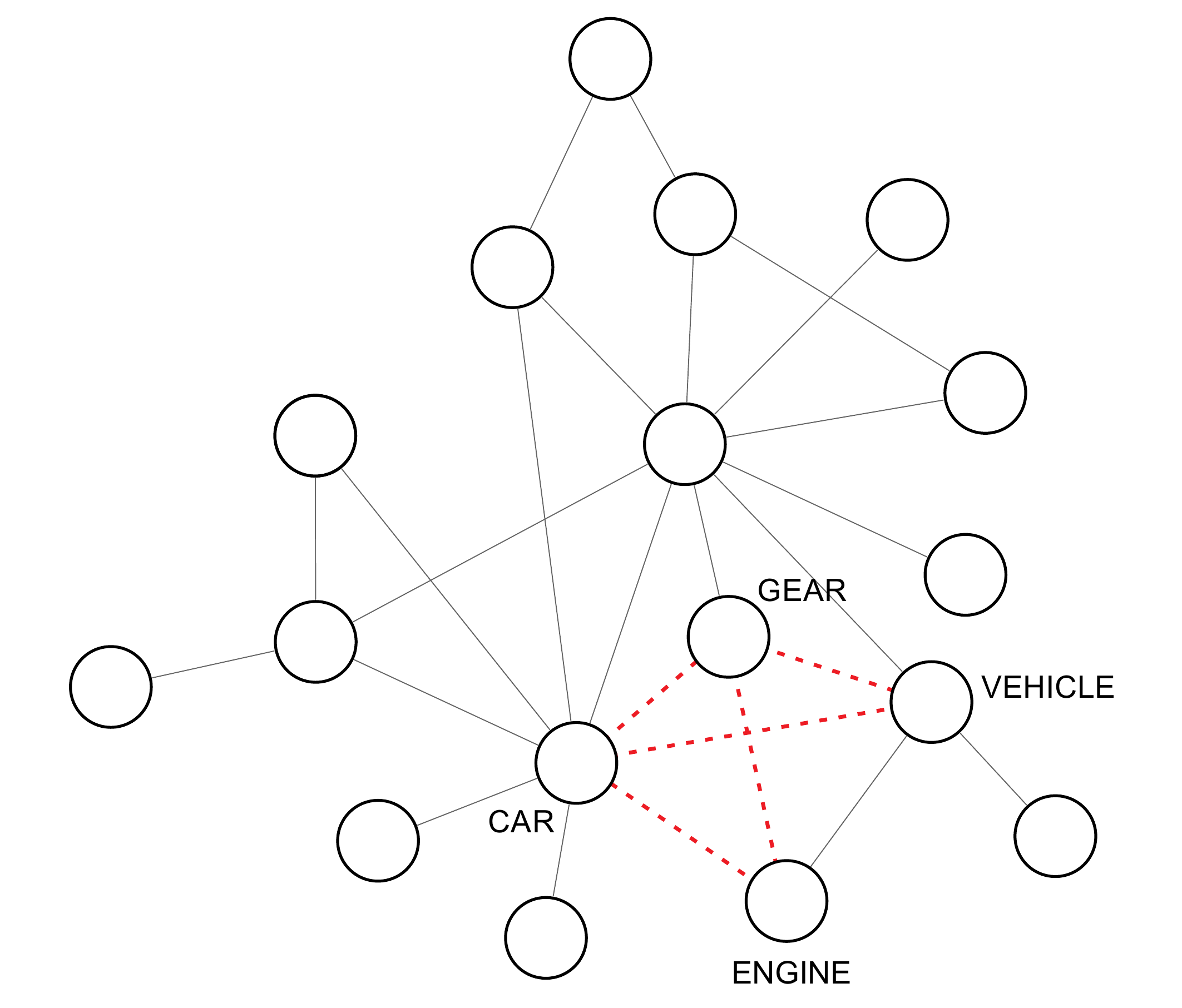}
    \caption{Example of a enriched word co-occurrence network created for a text. In this model, after the removal of stopwords, the remaining words are linked whenever they appear in the same context. In the proposed network representation, ``virtual'' edges are included whenever two nodes (words) are semantically related. In this example, virtual edges are those represented by red dashed lines. Edges are included via embeddings similarity. The quantity of included edges is a parameter to be chosen. }
   \label{fig:schematics}
\end{figure}

Such a drawback is addressed by including ``virtual'' edges connecting nodes. In other words, even if two words are not adjacent in the text, we include ``virtual'' edges to indicate that two distant words are semantically related. The inclusion of such virtual edges is illustrated in Figure \ref{fig:schematics}. In order to measure the semantical similarity between two concepts, we use the concept of \emph{word embeddings}~\cite{levy2015improving,rothe2015autoextend}. Thus, each word is represented using a vector representation encoding the semantical and contextual characteristics of the word. Several interesting properties have been obtained from distributed representation of words. One particular property encoded in the embeddings representation is the fact the semantical similarity between concepts is proportional to the similarity of vectors representing the words. Similarly to several other works, here we measure the similarity of the vectors via cosine similarity~\cite{nalisnick2016improving}.

The following strategies to create word embedding were considered in this paper:
\begin{enumerate}

    \item \emph{GloVe}: the Global Vectors (\emph{GloVe}) algorithm is an extension of the \emph{Word2vec} model~\cite{inproceedings} for efficient word vector learning~\cite{pennington2014glove}. This approach combines global statistics from matrix factorization techniques (such as latent semantic analysis) with context-based and predictive methods like \emph{Word2Vec}. This method is called as Global Vector method because the global corpus statistics are captured by \emph{GloVe}. Instead of using a window to define the local context, \emph{GloVe} constructs an explicit word-context matrix (or co-occurrence matrix) using statistics across the entire corpus. The final result is a learning model that oftentimes yields better word vector representations~\cite{pennington2014glove}.

    \item \emph{Word2Vec}: this is a predictive model that finds dense vector representations of words using a three-layer neural network with a single hidden layer~\cite{inproceedings}. It can be defined in a two-fold way: continuous bag-of-words and skip-gram model. In the latter, the model analyzes the words of a set of sentences (or corpus) and attempts to predict the neighbors of such words. For example, taking as reference the word ``Robin'', the model decides that ``Hood'' is more likely to follow the reference word than any other word. The vectors are obtained as follows: given the vocabulary (generated from all corpus words), the model trains a neural network with the sentences of the corpus. Then, for a given word, the probabilities that each word follows the reference word are obtained. Once the neural network is trained, the weights of the hidden layer are used as vectors of each corpus word.

    \item \emph{FastText}: this method is another extension of the \emph{Word2Vec} model~\cite{bojanowski2017enriching}. Unlike \emph{Word2Vec}, \emph{FastText} represents each word as a bag of character n-grams. Therefore, the neural network not only trains individual words, but also several n-grams of such words. The vector for a word is the sum of vectors obtained for the character n-grams composing the word. For example, the embedding obtained for the word ``computer'' with $n\leq3$ is the sum of the embeddings obtained for ``co'', ``com'', ``omp'', ``mpu'', ``put'', ``ute'', ``ter'' and ``er''. In this way, this method obtains improved representations for rare words, since n-grams composing rare words might be present in other words. The \emph{FastText} representation also allows the model to understand suffixes and prefixes. Another advantage of FastText is its efficiency to be trained in very large corpora.

\end{enumerate}

Concerning the thresholding process, we considered two main strategies. First, we used a global strategy: in addition to the co-occurrence links (continuous lines in Figure \ref{fig:schematics}), only ``virtual'' edges stronger than a given threshold are left in the network.
Thus only the most similar concepts are connected via virtual links. This strategy is hereafter referred to as \emph{global} strategy. Unfortunately, this method may introduce an undesired bias towards \emph{hubs}~\cite{serrano2009extracting}.
%and therefore other thresholding strategies should be considered.

To overcome the potential disadvantages of the global thresholding method, we  also considered a more refined thresholding approach that takes into account the local structure to decide whether a weighted link is statistically significant~\cite{serrano2009extracting}.  This method relies on the idea that the importance of an edge should be considered in the the context in which it appears. In other words, the relevance of an edge should be evaluated by analyzing the nodes connected to its ending points. Using the concept of \emph{disparity filter},
the method devised in~\cite{serrano2009extracting} defines a null model that quantifies the probability of a node to be connected to an edge with a given weight, based on its other connections. This probability is used to define the significance of the edge. The parameter that is used to measure the significance of an edge $e_{ij}$ is $\alpha_{ij}$, defined as:
\begin{equation} \label{eq:alpha}
    \alpha_{ij} = 1 - (k_i-1)\int_{0}^{\pi_{ij}}(1-x)^{k_i-2} dx ,
\end{equation}
\begin{equation} \label{eq:prob}
    \pi_{ij} = {w_{ij} \Bigg{(} \sum_{{ik}\,\in\,E}w_{ik}} \Bigg{)}^{-1},
\end{equation}
where $w_{ij}$ is the weight of the edge $e_{ij}$ and $k_i$ is the degree of the $i$-th node. The  obtained  network  corresponds to the set of nodes and edges obtained by  removing  all  edges  with $\alpha$ higher than the considered threshold. Note that while the similarity between co-occurrence links might be considered to compute $\alpha_{ij}$, only ``virtual'' edges (i.e. the dashed lines in Figure \ref{fig:schematics}) are eligible to be removed from the network in the filtering step. This strategy is hereafter referred to as \emph{local} strategy.

After co-occurrence networks are created and virtual edges are included, in the next step we used a characterization based on topological analysis. Because a global topological analysis is prone to variations in network size, we focused our analysis in the local characterization of complex networks. In a local topological analysis, we use as features the value of topological/dynamical measurements obtained for a set of words. In this case, we selected as feature the words occurring in all books of the dataset. For each word, we considered the following network measurements: degree, betweenness, clustering coefficient, average shortest path length, PageRank, concentric symmetry (at the second and third hierarchical level)~\cite{2015concentric} and accessibility~\cite{travenccolo2008accessibility,de2017knowledge} (at the second and third hierarchical level). We chose these measurements because all of them capture some particular linguistic feature of texts~\cite{liu2008complexity,liu2010language,amancio2012complex,yu2011statistical}.
After network measurements are extracted, they are used in machine learning algorithms. In our experiments, we considered Decision Trees~\footnote{This includes Random Forests.} (DT), nearest neighbors (kNN), Naive Bayes (NB) and Support Vector Machines (SVM).
We used some heuristics to optimize classifier parameters. Such techniques are described in the literature~\cite{rodriguez2019clustering}. The accuracy of the pattern recognition methods were evaluated using cross-validation~\cite{frank2004data}.

In summary, the methodology used in this paper encompasses the following steps:
\begin{enumerate}

    \item \emph{Network construction}: here texts are mapped into a co-occurrence networks. Some variations exists in the literature, however here we focused in the most usual variation, i.e. the possibility of considering or disregarding stopwords.
    A network with co-occurrence links is obtained after this step.

    \item \emph{Network enrichment}: in this step, the network is enriched with virtual edges established via similarity of word embeddings. After this step, we are given a complete network with weighted links. Virtually, any embedding technique could be used to gauge the similarity between nodes.

    \item \emph{Network filtering}: in order to eliminate spurious links included in the last step, the weakest edges are filtered. Two approaches were considered: a simple approach based on a global threshold and a local thresholding strategy that preserves network community structure. The outcome of this network filtering step is a network with two types of links: co-occurrence and virtual links (as shown in Figure \ref{fig:schematics}).

    \item \emph{Feature extraction}: In this step, topological and dynamical network features are extracted. Here, we do not discriminate co-occurrence from virtual edges to compute the network metrics.

    \item \emph{Pattern classification}: once features are extracted from complex networks, they are used in pattern classification methods. This might include supervised, unsupervised and semi-supervised classification. This framework is exemplified in the supervised scenario.

\end{enumerate}

The above framework is exemplified with the most common technique(s). It should be noted that the methods used, however, can be replaced by similar techniques. For example, the network construction could consider stopwords or even punctuation marks~\cite{kulig2017narrative}. Another possibility is the use of different strategies of thresholding.
While a systematic analysis of techniques and parameters is still required to reveal other potential advantages of the framework based on the addition of virtual edges, in this paper we provide a first analysis showing that virtual edges could be useful to improve the discriminability of texts modeled as complex networks.
%.... \textcolor{red}{dizer que pode mudar alguma coisa no proceso intermediario que o resultado pode melhorar}

Here we used a dataset compatible with datasets used recently in the literature (see e.g.~\cite{STANISZ2019301,segarra2015authorship,marinho2016authorship}). The objective of the studied stylometric task is to identify the authorship of an unknown document~\cite{basile2008example}.
All data and some statistics of each book are shown in the Supplementary Information.

\section{Results and Discussion}

In Section \ref{sec:perf}, we probe whether the inclusion of virtual edges is able to improve the performance of the traditional co-occurrence network-based classification in a usual stylometry task. While the focus of this paper is not to perform a systematic analysis of different methods comprising the adopted network, we consider two variations in the adopted methodology. In Section \ref{sec:stopthre}, we consider the use of stopwords and the adoption of a local thresholding process to establish different criteria to create new virtual edges.

%applied before the creation of the co-occurrence network affects the performance of the results. We also analyze . \textcolor{red}{In Section \ref{sec:top}, we probe the effect of including of additional edges on the centrality of nodes. Na outra secao mostrar o efeito de stopwords e metodo de filtragem}.
%
%In other words, we investigate whether virtual edges change the relative importance of words in the text.

\subsection{Performance analysis} \label{sec:perf}

\begin{figure}[h]
    \centering
    \includegraphics[width=0.75\textwidth]{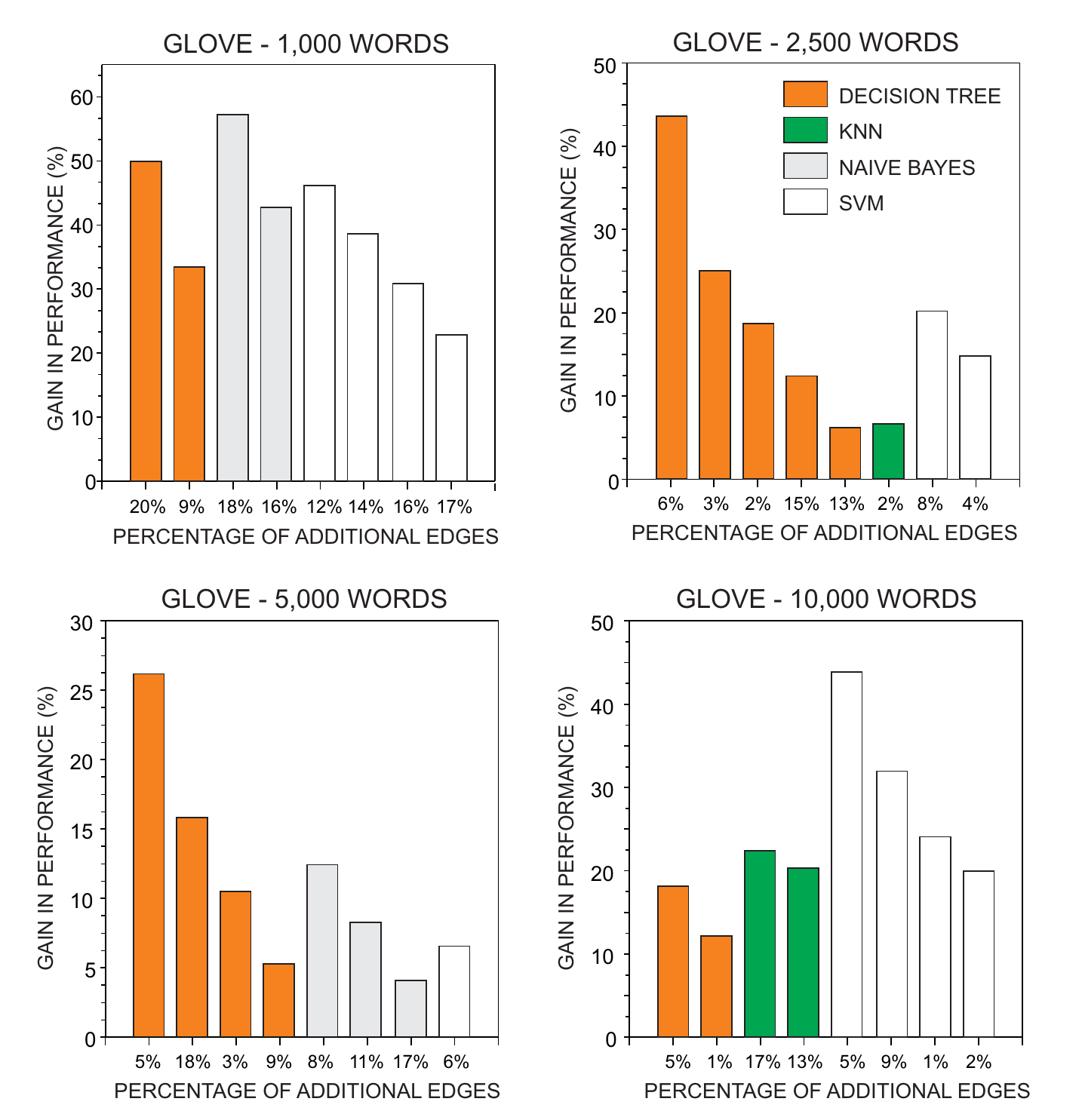}
    \caption{Gain in performance when considering additional virtual edges created using \emph{GloVe} as embedding method. Each sub-panel shows the results obtained for distinct values of text length. In this case, the highest improvements in performance tends to occur in the shortest documents.}
   \label{fig:glove}
\end{figure}

In Figure \ref{fig:glove}, we show some of the improvements in performance obtained when including a fixed amount of virtual edges using \emph{GloVe} as embedding method. In each subpanel, we show the relative improvement in performance obtained as a function of the fraction of additional edges.  In this section, we considered the traditional co-occurrence as starting point. In other words, the network construction disregarded stopwords. The list of stopwords considered in this paper is available in the Supplementary Information. We also considered the global approach to filter edges.

The relative improvement in performance is given by $\Gamma_+{(p)}/\Gamma_0$, where $\Gamma_+{(p)}$ is the accuracy rate obtained when $p\%$ additional edges are included and $\Gamma_0 = \Gamma_+{(p=0)}$, i.e. $\Gamma_0$ is the accuracy rate measured from the traditional co-occurrence model. We only show the highest relative improvements in performance for each classifier. In our analysis, we considered also samples of text with distinct length, since the performance of network-based methods is sensitive to text length~\cite{amancio2015probing}. In this figure, we considered samples comprising $w=\{1.0, 2.5, 5.0, 10.0\}$ thousand words.

The results obtained for \emph{GloVe} show that the highest relative improvements in performance occur for decision trees. This is apparent specially for the shortest samples. For $w=1,000$ words, the decision tree accuracy is enhanced by a factor of almost 50\% when $p=20\%$. An excellent gain in performance is also observed for both Naive Bayes and SVM classifiers, when $p=18\%$ and $p=12\%$, respectively. When $w=2,500$ words, the highest improvements was observed for the decision tree algorithm. A minor improvement was observed for the kNN method.  A similar behavior occurred for $w=5,000$ words. Interestingly, SVM seems to benefit from the use of additional edges when larger documents are considered. When only 5\% virtual edges are included, the relative gain in performance is about 45\%.

\begin{figure}[h]
    \centering
    \includegraphics[width=0.75\textwidth]{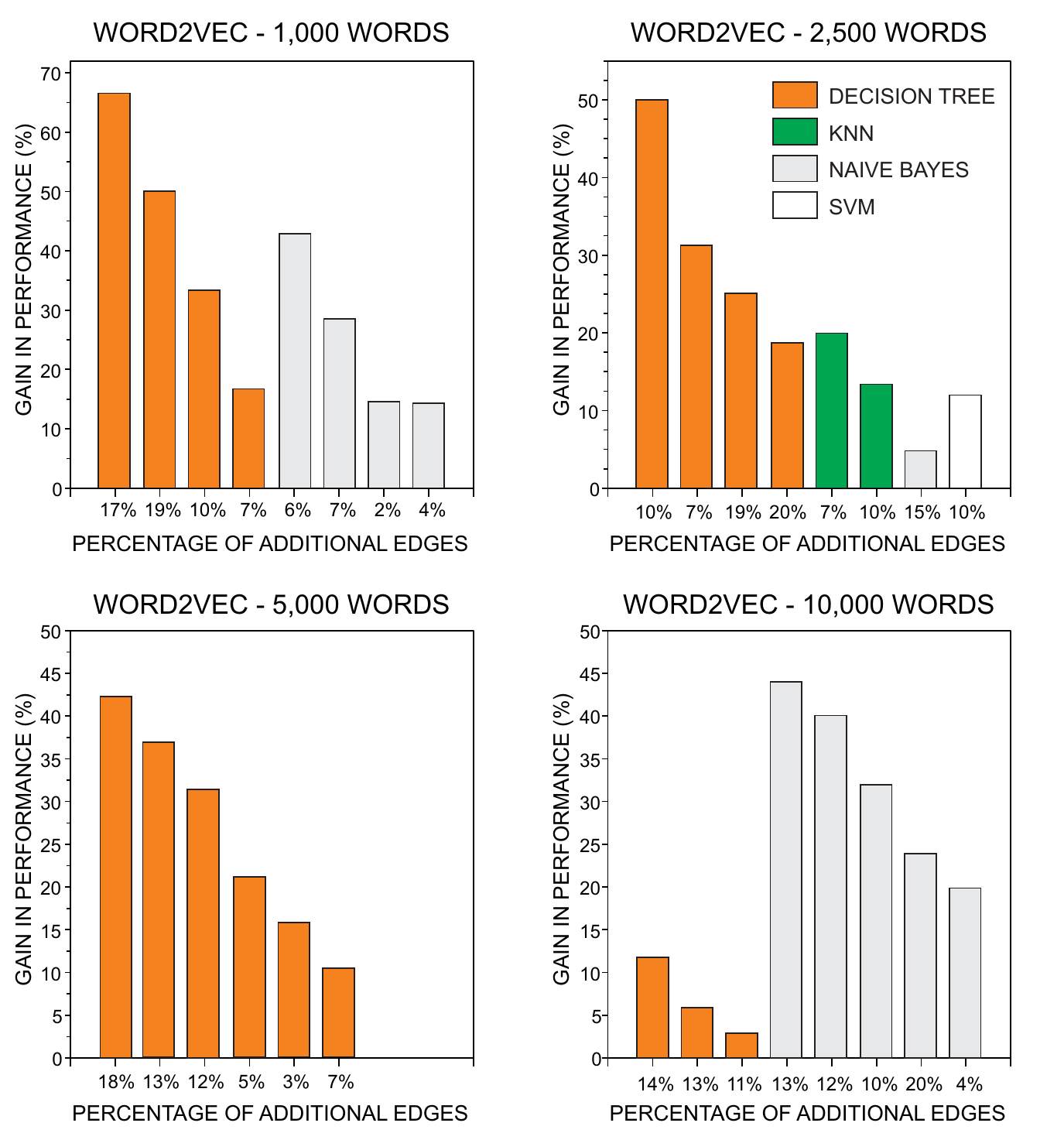}
    \caption{Gain in performance when considering additional virtual edges created using \emph{Word2vec} as embedding method. Each sub-panel shows the results obtained for distinct values of text length. %In this case, the highest improvements in performance tends to occur in shortest documents.
    }
   \label{fig:w2vec}
\end{figure}

The relative gain in performance obtained for \emph{Word2vec} is shown in Figure \ref{fig:w2vec}. Overall, once again decision trees obtained the highest gain in performance when short texts are considered. Similar to the analysis based on the \emph{GloVe} method, the gain for kNN is low when compared to the benefit received by other methods. Here, a considerable gain for SVM in only clear for $w=2,500$ and $p=10\%$. When large texts are considered, Naive Bayes obtained the largest gain in performance.

\begin{figure}[h]
    \centering
    \includegraphics[width=0.75\textwidth]{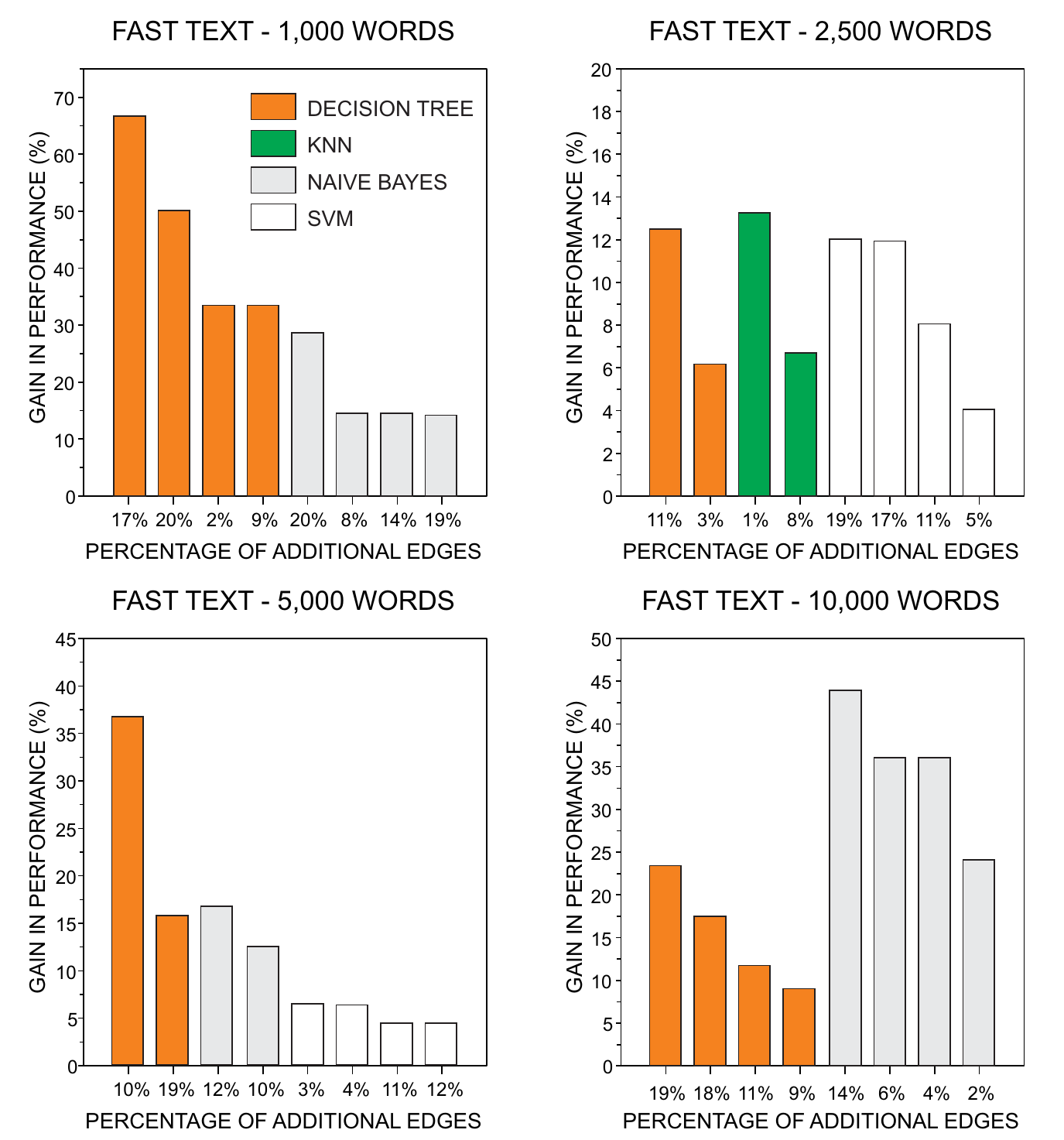}
    \caption{Gain in performance when considering additional virtual edges created using \emph{FastText} as embedding method. Each sub-panel shows the results obtained for distinct value of text length. %In this case, the highest improvements in performance tends to occur in shortest documents..
    }
   \label{fig:fasttext}
\end{figure}

Finally, the relative gain in performance obtained for \emph{FastText} is shown in Figure \ref{fig:fasttext}. The prominent role of virtual edges in decision tree algorithm in the classification of short texts once again is evident. Conversely, the classification of large documents using virtual edges mostly benefit the classification based on the Naive Bayes classifier. Similarly to the results observed for \emph{Glove} and \emph{Word2vec}, the gain in performance obtained for kNN is low compared when compared to other methods.

While Figures \ref{fig:glove} -- \ref{fig:fasttext} show the relative behavior in the accuracy, it still interesting to observe the absolute accuracy rate obtained with the classifiers. In Table \ref{table:gloves}, we show the best accuracy rate (i.e. $\max \Gamma_+ = \max_p \Gamma_+(p)$) for \emph{GloVe}. We also show the average difference in performance ($\langle \Gamma_+ - \Gamma_0 \rangle$) and the total number of cases in which an improvement in performance was observed ($N_+$). $N_+$ ranges in the interval $0 \leq N_+ \leq 20$. Table \ref{table:gloves} summarizes the results obtained for $w = \{1.0, 5.0, 10.0\}$ thousand words. \textcolor{black}{Additional results for other text length are available in Tables \ref{tabsi:glove-nostop}--\ref{tabsi:fast-nostop} of the Supplementary Information.}

In very short texts, despite the low accuracy rates, an improvement can be observed in all classifiers. The best results was obtained with SVM when virtual edges were included. For $w=5,000$ words, the inclusion of new edges has no positive effect on both kNN and Naive Bayes algorithms. On the other hand, once again SVM could be improved, yielding an optimized performance. For $w=10,000$ words, SVM could not be improved. However, even without improvement it yielded the maximum accuracy rate. The Naive Bayes algorithm, in average, could be improved by a margin of about 10\%.

%\singlespacing
\begin{table}[!htbp] \centering
   \caption{Statistics of performance obtained with \emph{GloVe} for different text lengths. Additional results considering other text lengths are shown in the Supplementary Information. $\Gamma_0$ is the the accuracy rate obtained with the traditional co-occurrence model and $max\ \Gamma_+$ is the highest accuracy rate considering different number of additional virtual edges. $\langle \Gamma_+ - \Gamma_0 \rangle$ is the average absolute improvement in performance, $\langle \Gamma_+ / \Gamma_0 \rangle$ is the average relative improvement in performance and $N_+$ is the total number of cases in which an improvement in performance was observed. In total we considered 20 different cases, which corresponds to the addition of $p=1\%$, $2\%$ $\ldots$ $20\%$ additional virtual edges. The best result for each document length is highlighted.
   %
   %Table for GloVe (1 a 20 c/normalizacao) s/stopwords.
   }
  \label{table:gloves}
  \footnotesize
\begin{tabular}{@{\extracolsep{5pt}}lcccc}
\\[-1.8ex]\hline
\hline \\[-1.8ex]
 & \multicolumn{4}{c}{\textit{1,000 words}} \\
\cline{2-5}
\\[-1.8ex] & DT & KNN & NB & SVM \\
%\\[-1.8ex] & (1) & (2)\\
\hline \\[-1.8ex]
$\Gamma_0$ & $15.38\%$ & $8.97\%$ & $8.97\%$ & $14.10\%$ \\
$\max\ \Gamma_+$ & ${\bf 16.67\%}$ & $10.26\%$ & $11.54\%$ & ${\bf 16.67\%}$ \\
$\langle \Gamma_+ - \Gamma_0\rangle$ & $1.29$ & $1.29$ & $1.61$ &$2.57$\\
$\langle \Gamma_+ / \Gamma_0 \rangle$ & $1.08$ & $1.14$ & $1.18$ & $1.18$\\
$N_+$ & $3$ & $3$ & $8$ & $2$\\
\hline \\[-1.8ex]

& \multicolumn{4}{c}{\textit{5,000 words}} \\
\cline{2-5}
$\Gamma_0$ & $24.36\%$  & $43.59\%$  & $30.77\%$ & $58.97\%$ \\
$\max\ \Gamma_+$ & $34.62\%$ & $-$ & $-$ & ${\bf 61.54\%}$ \\
$\langle \Gamma_+ - \Gamma_0\rangle$ & $6.70$ & $-$ & $-$ & $2.57$\\
$\langle \Gamma_+ / \Gamma_0 \rangle$ & $1.27$ & $-$ & $-$ & $1.04$\\
$N_+$ & $18$ & $0$ & $0$ & $1$\\
\hline \\[-1.8ex]

& \multicolumn{4}{c}{\textit{10,000 words}} \\
\cline{2-5}
$\Gamma_0$ & $42.31\%$  & $62.82\%$  & $32.05\%$ & ${\bf 85.90\%}$ \\
$\max\ \Gamma_+$ & $48.72\%$ & $74.36\%$ & $46.15\%$ & $-$ \\
$\langle \Gamma_+ - \Gamma_0\rangle$ & $2.84$ & $5.06$ & $9.68$ & $-$\\
$\langle \Gamma_+ / \Gamma_0 \rangle$ & $1.07$ & $1.08$ & $1.30$ & $-$\\
$N_+$ & $14$ & $18$ & $20$ & $0$\\
%\hline \\[-1.8ex]

\hline
\hline \\[-1.8ex]
\end{tabular}
\end{table}

%\doublespacing

The results obtained for \emph{Word2vec} are summarized in Table \ref{tabsi:w2vec-nostop} of the Supplementary Information. Considering short documents ($w=1,000$ words), here the best results occurs only with the decision tree method combined with enriched networks. Differently from the \emph{GloVe} approach, SVM does not yield the best results. Nonetheless, the highest accuracy across all classifiers and values of $p$ is the same. For larger documents ($w=5,000$ and $w=10,000$ words), no significant difference in performance between \emph{Word2vec} and \emph{GloVe} is apparent.

The results obtained for \emph{FastText} are shown in Table \ref{tab:fast}. In short texts, only kNN and Naive Bayes have their performance improved with virtual edges. However, none of the optimized results for these classifiers outperformed SVM applied to the traditional co-occurrence model. Conversely, when $w=5,000$ words, the optimized results are obtained with virtual edges in the SVM classifier. Apart from kNN, the enriched networks improved the traditional approach in all classifiers. For large chunks of texts ($w=10,000$), once again the approach based on SVM and virtual edges yielded optimized results. All classifiers benefited from the inclusion of additional edges. Remarkably, Naive Bayes improved by a margin of about $13\%$.

%\singlespacing
\begin{table}[!htbp] \centering
   \caption{Statistics of performance obtained with \emph{FastText} for different text lengths. Additional results considering other text lengths are shown in the Supplementary Information. $\Gamma_0$ is the the accuracy rate obtained with the traditional co-occurrence model and $\max\Gamma_+$ is the highest accuracy rate considering different number of additional virtual edges. $\langle \Gamma_+ - \Gamma_0 \rangle$ is the average absolute improvement in performance, $\langle \Gamma_+ / \Gamma_0 \rangle$ is the average relative improvement in performance and $N_+$ is the total number of cases in which an improvement in performance was observed. In total we considered 20 different cases, which corresponds to the addition of $p=1\%$, $2\%$ $\ldots$ $20\%$ additional virtual edges. The best result for each document length is highlighted.}
  \label{tab:fast}
  \footnotesize
\begin{tabular}{@{\extracolsep{5pt}}lcccc}
\\[-1.8ex]\hline
\hline \\[-1.8ex]
 & \multicolumn{4}{c}{\textit{1,000 words}} \\
\cline{2-5}
\\[-1.8ex] & DT & KNN & NB & SVM \\
%\\[-1.8ex] & (1) & (2)\\
\hline \\[-1.8ex]
$\Gamma_0$ & ${\bf 15.38\%}$ & $8.97\%$ & $8.97\%$ & ${14.10}\%$ \\
$\max \Gamma_+$ & $-$ & $10.26\%$ & $11.54\%$ & $-$ \\
$\langle \Gamma_+ - \Gamma_0\rangle$ & $-$ & $1.29$ & $1.57$ &$-$\\
$\langle \Gamma_+ / \Gamma_0 \rangle$ & $-$ & $1.14$ & $1.18$ & $-$\\
$N_+$ & $0$ & $2$ & $9$ & $0$\\
\hline \\[-1.8ex]

& \multicolumn{4}{c}{\textit{5,000 words}} \\
\cline{2-5}
$\Gamma_0$ & $24.36\%$  & $43.59\%$  & $30.77\%$ & $58.97\%$ \\
$\max \Gamma_+$ & $33.33\%$ & $-$ & $35.90\%$ & ${\bf 62.82\%}$ \\
$\langle \Gamma_+ - \Gamma_0 \rangle$ & $3.33$ & $-$ & $2.96$ & $2.34$\\
$\langle \Gamma_+ / \Gamma_0 \rangle$ & $1.14$ & $-$ & $1.10$ & $1.04$\\
$N_+$ & $10$ & $0$ & $13$ & $11$\\
\hline \\[-1.8ex]

& \multicolumn{4}{c}{\textit{10,000 words}} \\
\cline{2-5}
$\Gamma_0$ & $42.31\%$  & $62.82\%$  & $32.05\%$ & $85.90\%$ \\
$\max \Gamma_+$ & $53.85\%$ & $76.92\%$ & $48.72\%$ & ${\bf 87.18\%}$ \\
$\langle \Gamma_+ - \Gamma_0 \rangle$ & $6.49$ & $9.17$ & $12.96$ & $1.28$\\
$\langle \Gamma_+ / \Gamma_0 \rangle$ & $1.15$ & $1.15$ & $1.40$ & $1.01$\\
$N_+$ & $54$ & $20$ & $20$ & $4$\\
%\hline \\[-1.8ex]

\hline
\hline \\[-1.8ex]
\end{tabular}
\end{table}

%\doublespacing

%\textcolor{red}{Resumo dos melhores resultados aqui???}

\subsection{Effects of considering stopwords and local thresholding} \label{sec:stopthre}

While in the previous section we focused our analysis in the traditional word co-occurrence model, here we probe if the idea of considering virtual edges can also yield optimized results in particular modifications of the framework described in the methodology. The first modification in the co-occurrence model is the use of \emph{stopwords}. While in semantical application of network language modeling stopwords are disregarded, in other application it can unravel interesting linguistic patterns~\cite{segarra2015authorship}. Here we analyzed the effect of using stopwords in enriched networks. We summarize the obtained results in Table \ref{tab:stop}. We only show the results obtained with SVM, as it yielded the best results in comparison to other classifiers. The accuracy rate for other classifiers is shown in the Supplementary Information.

\begin{table}[h]
\caption{\label{tab:stop}Performance analysis of the adopted framework when considering \emph{stopwords} in the construction of the networks. Only the best results obtained across all considered classifiers are shown. In this case, all optimized results were obtained with SVM. $\Gamma_0$ corresponds to the accuracy obtained with no virtual edges and $\max \Gamma_+$ is the best accuracy rate obtained when including virtual edges. For each text length, the highest accuracy rate is highlighted. A full list of results for each classifier is available in the Supplementary Information. }
\small
\centering
\begin{tabular}{ccccc}
\hline
\multirow{1}{*}{\bf Length}  & \multirow{2}{*}{$\Gamma_0$} & $\max \Gamma_+$ & $\max \Gamma_+$ & $\max \Gamma_+$ \\
{\bf (words)}                 &                     & (\emph{GloVe})  & (\emph{Word2vec}) & (\emph{FastText})  \\
\hline
1,000 & {\bf 29.49\%} & 29.49\% & 29.49\% & 29.49\% \\
1,500 & 37.18\% & 37.18\% & 37.18\% & {\bf 38.46\%} \\
2,000 & 30.77\% & 34.62\% & {\bf 35.90\%} & {\bf 35.90\%} \\
2,500 & 41.03\% & 48.72\% & {\bf 51.28\%} & 48.72\% \\
5,000 & 62.82\% & {\bf 65.38\%} & 64.10\% & {\bf 65.38\%} \\
10,000 & {\bf 88.46\%} & 88.46\% & 88.46\% & 88.46\% \\
\hline
\end{tabular}
\end{table}

The results in Table \ref{tab:stop} reveals that even when stopwords are considered in the original model, an improvement can be observed with the addition of virtual edges. However, the results show that the degree of improvement depends upon the text length. In very short texts ($w=1,000$), none of the embeddings strategy was able to improve the performance of the classification. For $w=1,500$, a minor improvement was observed with \emph{FastText}: the accuracy increased from $\Gamma_0 = 37.18\%$ to $38.46\%$. A larger improvement could be observed for $w=2,000$. Both \emph{Word2vec} and \emph{FastText} approaches allowed an increase of more than 5\% in performance. A gain higher than 10\% was observed for $w=2,500$ with \emph{Word2vec}. For larger pieces of texts, the gain is less expressive or absent. All in all, the results show that the use of virtual edges can also benefit the network approach based on stopwords. However, no significant improvement could be observed with very short and very large documents. The comparison of all three embedding methods showed that no method performed better than the others in all cases.

We also investigated if more informed thresholding strategies could provide better results. While the simple global thresholding approach might not be able to represent more complex structures, we also tested a more robust approach based on the local approach proposed by Serrano et al.~\cite{serrano2009extracting}. In Table \ref{tab:limiar}, we summarize the results obtained with this thresholding strategies. The table shows $\max \Gamma_+^{(L)} / \max \Gamma_+^{(G)}$,  where $\Gamma_+^{(L)}$ and $\Gamma_+^{(G)}$ are the accuracy obtained with the local and global thresholding strategy, respectively. The results were obtained with the SVM classifier, as it turned to be the most efficient classification method. We found that there is no gain in performance when the local strategy is used. In particular cases, the global strategy is considerably more efficient. This is the case e.g. when \emph{GloVe} is employed in texts with $w=1,500$ words. The performance of the global strategy is $12.2\%$ higher than the one obtained with the global method. A minor difference in performance was found in texts comprising $w=1,000$ words, yet the global strategy is still more efficient than the global one.

\begin{table}[h]
\caption{\label{tab:limiar}Comparison between the best results obtained via global and local thresholding. For each text length and embedding method, we show $\max \Gamma_+^{(L)} / \max \Gamma_+^{(G)}$, where $\Gamma_+^{(L)}$ and $\Gamma_+^{(G)}$ are the accuracy obtained with local and global thresholding strategy, respectively. We only show the results obtained with the SVM, since it turned out to be the classifier yielding the highest accuracy rates. These results point that the use of this local strategy in the filtering process does not improve the performance of the classification. }
\small
\centering
\begin{tabular}{cccc}
\hline
{\bf Length}  & {\bf GloVe} & {\bf Word2vec} & {\bf FastText} \\
\hline
1,000 & 1.026  & 1.026 & 1.079 \\
1,500 & 1.122  & 1.093 & 1.019\\
2,000 & 1.068  & 1.091 & 1.091\\
2,500 & 1.020  & 1.061 & 1.082\\
5,000 & 1.036  & 1.054 & 1.071\\
10,000 & 1.045 & 1.030 & 1.015\\
\hline
\end{tabular}
\end{table}

%\subsectrmance for the kNN classifier ocion{Topological analysis} \label{sec:top}

To summarize all results obtained in this study we show in Table \ref{tab:summ} the best results obtained for each text length. We also show the relative gain in performance with the proposed approach and the embedding technique yielding the best result. All optimized results were obtained with the use of stopwords, global thresholding strategy and SVM as classification algorithm. A significant gain is more evident for intermediary text lengths.

\begin{table}[h]
\caption{\label{tab:summ}Summary of best results obtained in this paper. For each document length we show the highest accuracy rate obtained, the relative gain obtained with the proposed approach and the embedding method yielding the highest accuracy rate: \emph{GloVe} (GL), \emph{Word2Vec} (W2V) or \emph{FastText} (FT). All the results below were obtained when stopwords were used and the SVM was used as classification method. }
\small
\centering
\begin{tabular}{cccc}
\hline
{\bf Length}  & {\bf Accuracy} & {\bf Gain} & {\bf Embedding} \\
\hline
1,000 & 29.49\%  & -- & -- \\
1,500 & 38.46\%  &  3.44\% & FT \\
2,000 & 35.90\%  & 16.67\% & W2V, FT \\
2,500 & 51.28\%  & 24.98\% & W2V\\
5,000 & 65.38\%  & 4.07\% & GL, FT\\
10,000 & 88.46\% & -- & --\\
\hline
\end{tabular}
\end{table}

\section{Conclusion}

Textual classification remains one of the most important facets of the Natural Language Processing area. Here we studied a family of classification methods, the word co-occurrence networks.
%simple model that links adjacent words after some textual pre-processing steps.
Despite this apparent simplicity, this model has been useful in several practical and theoretical scenarios. We proposed a modification of the traditional model by establishing virtual edges to connect nodes that are semantically similar via word embeddings. The reasoning behind this strategy is the fact the similar words are not properly linked in the traditional model and, thus, important links might be overlooked if only adjacent words are linked.

Taking as reference task a stylometric problem, we showed -- as a proof of principle -- that the use of virtual edges might improve the discriminability of networks. When analyzing the best results for each text length, apart from very short and long texts, the proposed strategy yielded optimized results in all cases. The best classification performance was always obtained with the SVM classifier.
In addition, we found an improved performance when stopwords are used in the construction of the enriched co-occurrence networks. Finally, a simple global thresholding strategy was found to be more efficient than a local approach that preserves the community structure of the networks. Because complex networks are usually combined with other strategies~\cite{amancio2015complex,santos2017enriching}, we believe that the proposed
could be used in combination with other methods to improve the classification performance of other text classification tasks.
%\textcolor{red}{Dizer aqui os principais resultados....}

Our findings paves the way for research in several new directions. While we probed the effectiveness of virtual edges in a specific text classification task, we could extend this approach for general classification tasks. A systematic comparison of embeddings techniques could also be performed to include other recent techniques~\cite{devlin2018bert,yang2019xlnet}. We could also identify other relevant techniques to create virtual edges, allowing thus the use of the methodology in other networked systems other than texts. For example, a network could be enriched with embeddings obtained from graph embeddings techniques. A simpler approach could also consider link prediction~\cite{liben2007link} to create virtual edges. Finally, other interesting family of studies concerns the discrimination between co-occurrence and virtual edges, possibly by creating novel network measurements considering heterogeneous links.
%
%Finally, we intend to combine this strategies with others, in other to improve the state-of-the art algorithms.

%- idea of virtual edges could be extend to graph embeddings?
%- Could we use no embeddings at all, but link prediction?

%- Measures taking into account real and virtual edges?
%- other applications?
%- punctuation marks?

%- Study this aspect in other languages
%- Study the choice of best amount of edges before training

%- A systematic analysis of other embeddings techniques (BERT?), including a comparison between them in several other tasks

\section*{Acknowledgments}

The authors acknowledge financial support from FAPESP (Grant no.
16/19069-9), CNPq-Brazil (Grant no. 304026/2018-2). This study was financed in part by the Coordenação de Aperfeiçoamento de Pessoal de Nível Superior - Brasil (CAPES) - Finance Code 001.

\newpage

\bibliographystyle{ieeetr}
\bibliographystyle{abbrv}
%\bibliography{refs}

\newpage

\section*{Supplementary Information}

\subsection{Stopwords}

The following words were considered as stopwords in our analysis: all, just, don't, being, over, both, through, yourselves, its, before, o, don, hadn, herself, ll, had, should, to, only, won, under, ours,has, should've, haven't, do, them, his,
very, you've, they, not,  during, now, him, nor, wasn't,
d, did, didn, this, she, each, further, won't,
where, mustn't, isn't, few, because, you'd, doing,
some, hasn, hasn't, are, our, ourselves, out, what,
for, needn't, below, re, does, shouldn't, above, between, mustn, t, be, we, who, mightn't, doesn't, were, here, shouldn, hers, aren't,  by, on, about, couldn, of, wouldn't,
against, s, isn, or, own, into, yourself, down, hadn't, mightn, couldn't, wasn, your, you're, from, her, their, aren, it's, there, been, whom, too, wouldn, themselves, weren, was, until,
more, himself, that,  didn't, but, that'll, with, than, those, he, me, myself, ma, weren't, these, up, will, while, ain, can, theirs, my, and, ve, then, is, am, it, doesn, an, as, itself, at, have, in, any, if, again, no, when, same, how, other, which, you, shan't, shan, needn, haven, after, most, such, why, a, off i, m, yours, you'll, so, y, she's, the, having, once.

\subsection {List of books}

The list of books is shown in Tables \ref{tab:dataparte1} and \ref{tab:dataparte2}. For each book we show the respective authors (Aut.) and the following quantities: total number of words ($N_W$), total number of sentences ($N_S$), total number of paragraphs ($N_P$) and the average sentence length ($\langle S_L \rangle$), measured in number of words. The following authors were considered: Hector Hugh (HH), Thomas Hardy (TH), Daniel Defoe (DD), Allan Poe (AP), Bram Stoker (BS), Mark Twain (MT), Charles Dickens (CD), Pelham Grenville (PG), Charles Darwin (CD), Arthur Doyle (AD), George Eliot (GE), Jane Austen (JA), and Joseph Conrad (JC). %\textcolor{red}{completar aqui}

%\renewcommand{\tablename}{Table S}

%\makeatletter
%\renewcommand{\tablename}{Table S\@gobble}
%\makeatother

\setcounter{table}{0}
\renewcommand{\thetable}{S\arabic{table}}

\begin{table}[]
\caption{\label{tab:dataparte1}List of books used in the dataset comprising books written in English (first part). For each book we show the respective authors (Aut.) and the following quantities: total number of words ($N_W$), total number of sentences ($N_S$), total number of paragraphs ($N_P$) and the average sentence length ($\langle S_L \rangle$), in number of words. }
\small
\begin{tabular}{clcccc}
\hline
\textbf{Aut.}  & \textbf{Title (Publication Year)} & $N_W$ & $N_S$ & $N_P$ & $\langle S_L \rangle$  \\
\hline
HH  & The Toys of Peace (1919)& 67,734 & 2,045 & 1,109 & 33.12 \\
HH  & The Unbearable Bassington (1912) & 54,898 & 1,566 & 711 & 35.06 \\
HH  & Beasts and Super-Beasts (1914) & 73,944 & 2,289 & 1,354 & 32.30 \\
HH  & When William Came (1913) & 57,964 & 2,094 & 705 & 27.68 \\
HH  & The Rise of the Russian Empire (1900) & 133,859 & 3,376 & 807 & 39.65 \\
HH  & The Chronicles of Clovis (1912) & 61,176 & 2,467 & 1,051 & 24.80 \\
TH  & A Pair of Blue Eyes (1873) & 160,026 & 6,175 & 3,740 & 25.92 \\
TH  & A Changed Man (1913) & 103,500 & 5,093 & 1,845 & 20.32 \\
TH  & Far from the Madding Crowd (1874) & 166,225 & 8,898 & 172 & 18.68 \\
TH  & The Return of the Native (1878) & 169,820 & 6,518 & 3,485 & 26.05 \\
TH  & The Hand of Ethelberta $\ldots$ (1876) & 167,341 & 7,805 & 3,144 & 21.44 \\
TH  & Jude the Obscure (1895) & 176,298 & 9,294 & 3,622 & 18.97 \\
DD  & Memoirs of a Cavalier (1720) & 124,068 & 2,954 & 1,041 & 42.00 \\
DD  & Colonel Jack (1722) & 169,892 & 4,385 & 1,583 & 38.74 \\
DD  & The Fortunate Mistress (1724) & 190,768 & 3,860 & 1,554 & 49.42 \\
DD  & The Life, Adventures \& Piracies $\ldots$ (1720) & 131,701 & 2,480 & 1,060 & 53.11\\
DD  & The Fortunes and Misfortunes $\ldots$ (1722) & 159,512 & 3,556 & 1,281 & 44.86 \\
DD  & The Farther Adventures of Robinson $\ldots$ (1719) & 138,328 & 2,204 & 739 & 62.76 \\
AP  & The Works of Edgar Allan Poe - V1 (1850) & 106,902 & 3,516 & 867 & 30.40 \\
AP  & The Works of Edgar Allan Poe - V2 (1859) & 113,124 & 3,791 & 1,002 & 29.84 \\
AP  & The Works of Edgar Allan Poe - V3 (1859) & 115,605 & 3,586 & 680 & 32.24 \\
AP  & The Works of Edgar Allan Poe - V4 (1859) & 105,246 & 3,829 & 1,106 & 27.49 \\
AP  & The Works of Edgar Allan Poe - V5 (1859) & 89,002 & 3,099 & 1,339 & 28.72 \\
AP  & The Narrative of Arthur Gordon $\ldots$ (1838) & 81,305 & 2,476 & 380 & 32.84 \\
BS  & The Lady of the Shroud (1909) & 147,990 & 6,325 & 1,469 & 23.40 \\
BS  & The Mystery of the Sea (1902) & 180,282 & 7,811 & 1,914 & 23.08 \\
BS  & The Jewel of Seven Stars (1903) & 103,961 & 4,863 & 1,232 & 21.38 \\
BS  & The Lair of the White Worm (1911) & 65,032 & 3,125 & 898 & 20.81 \\
BS  & The Man (1905) & 121,726 & 7,049 & 1,871 & 17.27 \\
BS  & Dracula's Guest (1914) & 65,723 & 2,863 & 749 & 22.96 \\
MT  & Following the Equator: A Journey $\ldots$ (1897) & 219,900 & 8,609 & 2,305 & 25.54 \\
MT  & Life on the Mississippi (1883) & 170,776 & 6,745 & 2,095 & 25.32 \\
MT  & The Prince and the Pauper (1881) & 85,398 & 2,687 & 1,636 & 31.78 \\
MT  & The Innocents Abroad (1869) & 224,169 & 8,056 & 1,985 & 27.83 \\
MT  & Adventures of Huckleberry Finn (1884) & 136,841 & 5,798 & 2,225 & 23.60 \\
MT  & The Adventures of Tom Sawyer (1876) & 87,953 & 3,679 & 2,100 & 23.91 \\
CD  & Oliver Twist (1837) & 195,337 & 9,205 & 3,961 & 21.22 \\
CD  & David Copperfield (1849) & 443,613 & 14,952 & 7,190 & 29.67 \\
CD  & The Mystery of Edwin Drood (1870) & 118,250 & 3,888 & 2,527 & 30.41 \\
\hline
\end{tabular}
\end{table}

\begin{table}[]
\caption{\label{tab:dataparte2}List of books used in the dataset comprising books written in English (second part). For each book we show the respective authors (Aut.) and the following quantities: total number of words ($N_W$), total number of sentences ($N_S$), total number of paragraphs ($N_P$) and the average sentence length ($\langle S_L \rangle$), in number of words.}
\small
\begin{tabular}{clcccc}
\hline
\textbf{Aut.}  & \textbf{Title (Publication Year)} & $N_W$ & $N_S$ & $N_P$ & $\langle S_L \rangle$  \\ \hline
CD  & Barnaby Rudge: A Tale of the $\ldots$ (1841) & 315,297 & 9,345 & 4,699 & 33.74 \\
CD & The Pickwick Papers (1836) & 387,770 & 11,543 & 8,112 & 33.59 \\
CD & A Tale of Two Cities (1859) & 165,555 & 5,689 & 3,327 & 29.10 \\
PG & Right Ho, Jeeves (1934) & 93,293 & 6,374 & 3,228 & 14.64 \\
PG & My Man Jeeves (1919) & 64,323 & 4,670 & 1,951 & 13.77 \\
PG & The Clicking of Cuthbert (1922) & 74,581 & 4,788 & 1,888 & 15.58 \\
PG & The Man with Two Left Feet (1917) & 85,686 & 5,719 & 2,188 & 14.98 \\
PG & The Adventures of Sally (1921) & 97,402 & 4,869 & 2,367 & 20.00 \\
PG & Tales of St. Austin's (1903) & 61,700 & 4,014 & 1,355 & 15.37 \\
CD & Geological Observations on South America (1846) & 151,983 & 4,432 & 826 & 34.29 \\
CD & Geological Observations on the Volcanic Islands (1844) & 64,455 & 1,837 & 303 & 35.09 \\         CD & The Structure and Distribution of Coral Reefs (1842) & 99,069 & 2,554 & 435 & 38.79 \\
CD & The Different Forms of Flowers on $\ldots$ (1877) & 116,526 & 4,670 & 1,166 & 24.95 \\
CD & The Expression of the Emotions in $\ldots$ (1872) & 113,960 & 3,394 & 628 & 33.58 \\
CD & On the origin of species (1859) & 176,250 & 4,783 & 1,367 & 36.85 \\
AD & The Adventures of Sherlock Holmes (1892) & 125,740 & 6,831 & 2,541 & 18.41 \\
AD & The Refugees (1893) & 147,193 & 7,816 & 3,112 & 18.83 \\
AD & The Lost World (1912) & 89,274 & 4,459 & 1,237 & 20.02 \\
AD & The Exploits of Brigadier Gerard (1896) & 86,314 & 4,027 & 1,313 & 21.43 \\
AD & The Valley of Fear (1915) & 70,557 & 3,062 & 1,538 & 23.04 \\
AD & Micah Clarke (1889) & 210,234 & 7,244 & 2,518 & 29.02 \\
GE & The Mill on the Floss (1860) & 250,748 & 8,877 & 3,208 & 28.25 \\
GE & Adam Bede (1859) & 255,007 & 7,131 & 2,572 & 35.76 \\
GE & Romola (1862) & 264,275 & 9,093 & 2,906 & 29.06 \\
GE & Daniel Deronda (1876) & 362,291 & 14,350 & 4,385 & 25.25 \\
GE & Middlemarch (1871) & 373,085 & 14,885 & 4,797 & 25.07 \\
GE & Felix Holt, the Radical (1866) & 214,122 & 8,072 & 2,571 & 26.53 \\
JA & Mansfield Park (1814) & 185,880 & 5,722 & 1,840 & 32.49 \\
JA & Sense and Sensibility (1811) & 141,356 & 4,835 & 1,864 & 29.24 \\
JA & Northanger Abbey (1817) & 91,042 & 2,748 & 1,056 & 33.13 \\
JA & Persuasion (1818) & 97,854 & 3,653 & 1,035 & 26.79 \\
JA & Emma (1816) & 190,481 & 5,911 & 2,375 & 32.23 \\
JA & Pride and Prejudice (1813) & 142,455 & 4,671 & 2,126 & 30.50 \\
JC & Victory: An Island Tale (1915) & 142,609 & 7,314 & 2,730 & 19.50 \\
JC & Lord Jim (1900) & 157,066 & 8,028 & 700 & 19.56 \\
JC & Chance: A Tale in Two Parts (1913) & 161,744 & 9,374 & 1,924 & 17.25 \\
JC & Nostromo: A Tale of the Seaboard (1904) & 200,945 & 8,834 & 2,251 & 22.75 \\
JC & Under Western Eyes (1911) & 135,160 & 7,118 & 2,373 & 18.99 \\
JC & An Outcast of the Islands (1896) & 128,575 & 7,917 & 1,648 & 16.24 \\
\hline
\end{tabular}
\end{table}

\newpage

\subsection{Additional results}

In this section we show additional results obtained for different text length. More specifically, we show the results obtained for \emph{GloVe}, \emph{Word2vec} and \emph{FastText} when \emph{stopwords} are either considered in the text or disregarded from the analysis.

%\singlespacing
\begin{table}[!htbp] \centering
   \caption{
   Statistics of performance obtained with \emph{GloVe}. The network construction phase \emph{disregarded} stopwords. $\Gamma_0$ is the the accuracy rate obtained with the traditional co-occurrence model and $max\ \Gamma_+$ is the highest accuracy rate considering different number of additional virtual edges. $\langle \Gamma_+ - \Gamma_0 \rangle$  and $\langle \Gamma_+ / \Gamma_0 \rangle$ are the average absolute and relative improvement in performance, respectively. $N_+$ is the total number of cases with an improvement in performance.
   }
  \label{tabsi:glove-nostop}
  \footnotesize
\begin{tabular}{@{\extracolsep{5pt}}lcccc}
\\[-1.8ex]\hline
\hline \\[-2.8ex]
 & \multicolumn{4}{c}{\textit{1,000 words}} \\
\cline{2-5}
\\[-1.8ex] & DT & KNN & NB & SVM \\
%\\[-1.8ex] & (1) & (2)\\
\hline \\[-2.8ex]
$\Gamma_0$ & $15.38\%$ & $8.97\%$ & $8.97\%$ & $14.10\%$ \\
$max\ \Gamma_+$ & $16.67\%$ & $10.26\%$ & $11.54\%$ & $16.67\%$ \\
$\langle \Gamma_+ - \Gamma_0 \rangle$ & $1.29$ & $1.29$ & $1.61$ &$2.57$\\
$\langle \Gamma_+ / \Gamma_0 \rangle$ & $1.08$ & $1.14$ & $1.18$ & $1.18$\\
$N_+$ & $3$ & $3$ & $8$ & $2$\\
\hline \\[-2.8ex]
 & \multicolumn{4}{c}{\textit{1,500 words}} \\
\cline{2-5}
$\Gamma_0$ & $7.69\%$  & $12.82\%$  & $20.51\%$ & $12.82\%$ \\
$max\ \Gamma_+$ & $19.23\%$ & $16.67\%$ & $24.36\%$ & $23.08\%$ \\
$\langle \Gamma_+ \rangle$ & $4.81$ & $2.45$ & $1.97$ & $6.21$\\
$\langle \Gamma_+ / \Gamma_0 \rangle$ & $1.63$ & $1.19$ & $1.10$ & $1.48$\\
$N_+$ & $20$ & $11$ & $13$ & $19$\\
\hline \\[-2.8ex]

 & \multicolumn{4}{c}{\textit{2,000 words}} \\
\cline{2-5}
$\Gamma_0$ & $14.10\%$  & $15.38\%$  & $14.10\%$ & $23.08\%$ \\
$max\ \Gamma_+$ & $23.08\%$ & $21.79\%$ & $23.08\%$ & $25.64\%$ \\
$\langle \Gamma_+ - \Gamma_0 \rangle$ & $4.95$ & $3.45$ & $4.43$ & $1.40$\\
$\langle \Gamma_+ / \Gamma_0 \rangle$ & $1.35$ & $1.23$ & $1.31$ & $1.06$\\
$N_+$ & $14$ & $20$ & $20$ & $11$\\
\hline \\[-2.8ex]

& \multicolumn{4}{c}{\textit{2,500 words}} \\
\cline{2-5}
$\Gamma_0$ & $20.51\%$  & $19.23\%$  & $26.92\%$ & $32.05\%$ \\
$max\ \Gamma_+$ & $29.49\%$ & $20.51\%$ & $-$ & $38.46\%$ \\
$\langle \Gamma_+ - \Gamma_0 \rangle$ & $4.32$ & $1.28$ & $-$ & $3.45$\\
$\langle \Gamma_+ / \Gamma_0 \rangle$ & $1.21$ & $1.07$ & $-$ & $1.11$\\
$N_+$ & $11$ & $2$ & $0$ & $13$\\
\hline \\[-2.8ex]

& \multicolumn{4}{c}{\textit{5,000 words}} \\
\cline{2-5}
$\Gamma_0$ & $24.36\%$  & $43.59\%$  & $30.77\%$ & $58.97\%$ \\
$max\ \Gamma_+$ & $30.77\%$ & $-$ & $34.62\%$ & $62.82\%$ \\
$\langle \Gamma_+ - \Gamma_0 \rangle$ & $2.99$ & $-$ & $2.84$ & $2.31$\\
$\langle \Gamma_+ / \Gamma_0 \rangle$ & $1.12$ & $-$ & $1.09$ & $1.04$\\
$N_+$ & $9$ & $0$ & $14$ & $15$\\
\hline \\[-2.8ex]

& \multicolumn{4}{c}{\textit{10,000 words}} \\
\cline{2-5}
$\Gamma_0$ & $42.31\%$  & $62.82\%$  & $32.05\%$ & $85.90\%$ \\
$max\ \Gamma_+$ & $50.00\%$ & $76.92\%$ & $46.15\%$ & $-$ \\
$\langle \Gamma_+ - \Gamma_0 \rangle$ & $4.49$ & $11.03$ & $9.04$ & $-$\\
$\langle \Gamma_+ / \Gamma_0 \rangle$ & $1.11$ & $1.18$ & $1.28$ & $-$\\
$N_+$ & $6$ & $20$ & $20$ & $0$\\
%\hline \\[-1.8ex]

\hline
\hline \\[-1.8ex]
\end{tabular}
\end{table}

%\singlespacing
\begin{table}[!htbp] \centering
   \caption{
   Statistics of performance obtained with \emph{word2vec}. The network construction phase \emph{disregarded} stopwords. $\Gamma_0$ is the the accuracy rate obtained with the traditional co-occurrence model and $max\ \Gamma_+$ is the highest accuracy rate considering different number of additional virtual edges. $\langle \Gamma_+ - \Gamma_0 \rangle$  and $\langle \Gamma_+ / \Gamma_0 \rangle$ are the average absolute and relative improvement in performance, respectively. $N_+$ is the total number of cases with an improvement in performance.
   }
  \label{tabsi:w2vec-nostop}
  \footnotesize
\begin{tabular}{@{\extracolsep{5pt}}lcccc}
\\[-1.8ex]\hline
\hline \\[-2.8ex]
 & \multicolumn{4}{c}{\textit{1,000 words}} \\
\cline{2-5}
\\[-1.8ex] & DT & KNN & NB & SVM \\
%\\[-1.8ex] & (1) & (2)\\
\hline \\[-1.8ex]
$\Gamma_0$ & $15.38\%$ & $8.97\%$ & $8.97\%$ & $14.10\%$ \\
$max\ \Gamma_+$ & $16.67\%$ & $10.26\%$ & $12.82\%$ & $-$ \\
$\langle \Gamma_+ - \Gamma_0 \rangle$ & $1.29$ & $1.29$ & $1.76$ &$-$\\
$\langle \Gamma_+ / \Gamma_0 \rangle$ & $1.08$ & $1.14$ & $1.20$ & $-$\\
$N_+$ & $1$ & $3$ & $11$ & $0$\\
\hline \\[-2.8ex]
 & \multicolumn{4}{c}{\textit{1,500 words}} \\
\cline{2-5}
$\Gamma_0$ & $7.69\%$  & $12.82\%$  & $20.51\%$ & $12.82\%$ \\
$max\ \Gamma_+$ & $15.38\%$ & $15.38\%$ & $23.08\%$ & $23.08\%$ \\
$\langle \Gamma_+ - \Gamma_0 \rangle$ & $3.67$ & $1.54$ & $1.56$ & $7.18$\\
$\langle \Gamma_+ / \Gamma_0 \rangle$ & $1.48$ & $1.12$ & $1.08$ & $1.56$\\
$N_+$ & $14$ & $5$ & $14$ & $20$\\
\hline \\[-2.8ex]

 & \multicolumn{4}{c}{\textit{2,000 words}} \\
\cline{2-5}
$\Gamma_0$ & $14.10\%$  & $15.38\%$  & $14.10\%$ & $23.08\%$ \\
$max\ \Gamma_+$ & $26.92\%$ & $29.49\%$ & $28.21\%$ & $26.92\%$ \\
$\langle \Gamma_+ - \Gamma_0 \rangle$ & $5.47$ & $9.88$ & $8.14$ & $1.86$\\
$\langle \Gamma_+ / \Gamma_0 \rangle$ & $1.39$ & $1.64$ & $1.58$ & $1.08$\\
$N_+$ & $8$ & $19$ & $17$ & $17$\\
\hline \\[-2.8ex]

& \multicolumn{4}{c}{\textit{2,500 words}} \\
\cline{2-5}
$\Gamma_0$ & $20.51\%$  & $19.23\%$  & $26.92\%$ & $32.05\%$ \\
$max\ \Gamma_+$ & $34.62\%$ & $20.51\%$ & $32.05\%$ & $38.46\%$ \\
$\langle \Gamma_+ - \Gamma_0 \rangle$ & $7.27$ & $1.28$ & $2.83$ & $3.85$\\
$\langle \Gamma_+ / \Gamma_0 \rangle$ & $1.35$ & $1.07$ & $1.10$ & $1.12$\\
$N_+$ & $18$ & $2$ & $15$ & $15$\\
\hline \\[-2.8ex]

& \multicolumn{4}{c}{\textit{5,000 words}} \\
\cline{2-5}
$\Gamma_0$ & $24.36\%$  & $43.59\%$  & $30.77\%$ & $58.97\%$ \\
$max\ \Gamma_+$ & $34.62\%$ & $-$ & $-$ & $61.54\%$ \\
$\langle \Gamma_+ - \Gamma_0 \rangle$ & $6.70$ & $-$ & $-$ & $2.57$\\
$\langle \Gamma_+ / \Gamma_0 \rangle$ & $1.27$ & $-$ & $-$ & $1.04$\\
$N_+$ & $18$ & $0$ & $0$ & $1$\\
\hline \\[-2.8ex]

& \multicolumn{4}{c}{\textit{10,000 words}} \\
\cline{2-5}
$\Gamma_0$ & $42.31\%$  & $62.82\%$  & $32.05\%$ & $85.90\%$ \\
$max\ \Gamma_+$ & $48.72\%$ & $74.36\%$ & $46.15\%$ & $-$ \\
$\langle \Gamma_+ - \Gamma_0 \rangle$ & $2.84$ & $5.06$ & $9.68$ & $-$\\
$\langle \Gamma_+ / \Gamma_0 \rangle$ & $1.07$ & $1.08$ & $1.30$ & $-$\\
$N_+$ & $14$ & $18$ & $20$ & $0$\\
%\hline \\[-1.8ex]

\hline
\hline \\[-1.8ex]
\end{tabular}
\end{table}

%\singlespacing
\begin{table}[!htbp] \centering
   \caption{Statistics of performance obtained with \emph{FastText}. The network construction phase \emph{disregarded} stopwords. $\Gamma_0$ is the the accuracy rate obtained with the traditional co-occurrence model and $max\ \Gamma_+$ is the highest accuracy rate considering different number of additional virtual edges. $\langle \Gamma_+ - \Gamma_0 \rangle$  and $\langle \Gamma_+ / \Gamma_0 \rangle$ are the average absolute and relative improvement in performance, respectively. $N_+$ is the total number of cases with an improvement in performance.}
  \label{tabsi:fast-nostop}
  \footnotesize
\begin{tabular}{@{\extracolsep{5pt}}lcccc}
\\[-1.8ex]\hline
\hline \\[-2.8ex]
 & \multicolumn{4}{c}{\textit{1,000 words}} \\
\cline{2-5}
\\[-1.8ex] & DT & KNN & NB & SVM \\
%\\[-1.8ex] & (1) & (2)\\
\hline \\[-1.8ex]
$\Gamma_0$ & $15.38\%$ & $8.97\%$ & $8.97\%$ & $14.10\%$ \\
$max\ \Gamma_+$ & $-$ & $10.26\%$ & $11.54\%$ & $-$ \\
$\langle \Gamma_+ - \Gamma_0 \rangle$ & $-$ & $1.29$ & $1.57$ &$-$\\
$\langle \Gamma_+ / \Gamma_0 \rangle$ & $-$ & $1.14$ & $1.18$ & $-$\\
$N_+$ & $0$ & $2$ & $9$ & $0$\\
\hline \\[-2.8ex]
 & \multicolumn{4}{c}{\textit{1,500 words}} \\
\cline{2-5}
$\Gamma_0$ & $7.69\%$  & $12.82\%$  & $20.51\%$ & $12.82\%$ \\
$max\ \Gamma_+$ & $15.38\%$ & $20.51\%$ & $24.36\%$ & $20.51\%$ \\
$\langle \Gamma_+ - \Gamma_0 \rangle$ & $3.55$ & $3.54$ & $2.41$ & $4.91$\\
$\langle \Gamma_+ / \Gamma_0 \rangle$ & $1.46$ & $1.28$ & $1.12$ & $1.38$\\
$N_+$ & $17$ & $17$ & $8$ & $18$\\
\hline \\[-2.8ex]

 & \multicolumn{4}{c}{\textit{2,000 words}} \\
\cline{2-5}
$\Gamma_0$ & $14.10\%$  & $15.38\%$  & $14.10\%$ & $23.08\%$ \\
$max\ \Gamma_+$ & $23.08\%$ & $21.79\%$ & $21.79\%$ & $30.77\%$ \\
$\langle \Gamma_+ - \Gamma_0 \rangle$ & $4.20$ & $3.98$ & $4.43$ & $4.45$\\
$\langle \Gamma_+ / \Gamma_0 \rangle$ & $1.30$ & $1.26$ & $1.31$ & $1.19$\\
$N_+$ & $11$ & $20$ & $20$ & $17$\\
\hline \\[-2.8ex]

& \multicolumn{4}{c}{\textit{2,500 words}} \\
\cline{2-5}
$\Gamma_0$ & $20.51\%$  & $19.23\%$  & $26.92\%$ & $32.05\%$ \\
$max\ \Gamma_+$ & $26.92\%$ & $21.79\%$ & $-$ & $35.90\%$ \\
$\langle \Gamma_+ - \Gamma_0 \rangle$ & $2.89$ & $1.71$ & $-$ & $2.33$\\
$\langle \Gamma_+ / \Gamma_0 \rangle$ & $1.14$ & $1.09$ & $-$ & $1.07$\\
$N_+$ & $4$ & $3$ & $0$ & $16$\\
\hline \\[-2.8ex]

& \multicolumn{4}{c}{\textit{5,000 words}} \\
\cline{2-5}
$\Gamma_0$ & $24.36\%$  & $43.59\%$  & $30.77\%$ & $58.97\%$ \\
$max\ \Gamma_+$ & $33.33\%$ & $-$ & $35.90\%$ & $62.82\%$ \\
$\langle \Gamma_+ - \Gamma_0 \rangle$ & $3.33$ & $-$ & $2.96$ & $2.34$\\
$\langle \Gamma_+ / \Gamma_0 \rangle$ & $1.14$ & $-$ & $1.10$ & $1.04$\\
$N_+$ & $10$ & $0$ & $13$ & $11$\\
\hline \\[-2.8ex]

& \multicolumn{4}{c}{\textit{10,000 words}} \\
\cline{2-5}
$\Gamma_0$ & $42.31\%$  & $62.82\%$  & $32.05\%$ & $85.90\%$ \\
$max\ \Gamma_+$ & $53.85\%$ & $76.92\%$ & $48.72\%$ & $87.18\%$ \\
$\langle \Gamma_+ - \Gamma_0 \rangle$ & $6.49$ & $9.17$ & $12.96$ & $1.28$\\
$\langle \Gamma_+ / \Gamma_0 \rangle$ & $1.15$ & $1.15$ & $1.40$ & $1.01$\\
$N_+$ & $54$ & $20$ & $20$ & $4$\\
%\hline \\[-1.8ex]

\hline
\hline \\[-1.8ex]
\end{tabular}
\end{table}

%\singlespacing
\begin{table}[!htbp] \centering
   \caption{Statistics of performance obtained with \emph{GloVe}. The network construction phase \emph{considered} stopwords. $\Gamma_0$ is the the accuracy rate obtained with the traditional co-occurrence model and $max\ \Gamma_+$ is the highest accuracy rate considering different number of additional virtual edges. $\langle \Gamma_+ - \Gamma_0 \rangle$  and $\langle \Gamma_+ / \Gamma_0 \rangle$ are the average absolute and relative improvement in performance, respectively. $N_+$ is the total number of cases with an improvement in performance.}
  \label{tabdetfin}
  \footnotesize
\begin{tabular}{@{\extracolsep{5pt}}lcccc}
\\[-1.8ex]\hline
\hline \\[-2.8ex]
 & \multicolumn{4}{c}{\textit{1,000 words}} \\
\cline{2-5}
\\[-1.8ex] & DT & KNN & NB & SVM \\
%\\[-1.8ex] & (1) & (2)\\
\hline \\[-1.8ex]
$\Gamma_0$ & $16.67\%$ & $17.95\%$ & $17.95\%$ & $29.49\%$ \\
$max\ \Gamma_+$ & $25.64\%$ & $20.51\%$ & $24.36\%$ & $-$ \\
$\langle \Gamma_+ - \Gamma_0 \rangle$ & $4.30$ & $1.60$ & $3.62$ &$-$\\
$\langle \Gamma_+ / \Gamma_0 \rangle$ & $1.26$ & $1.09$ & $1.20$ & $-$\\
$N_+$ & $14$ & $4$ & $17$ & $0$\\
\hline \\[-2.8ex]
 & \multicolumn{4}{c}{\textit{1,500 words}} \\
\cline{2-5}
$\Gamma_0$ & $23.08\%$  & $21.79\%$  & $30.77\%$ & $37.18\%$ \\
$max\ \Gamma_+$ & $-$ & $24.36\%$ & $37.18\%$ & $-$ \\
$\langle \Gamma_+ - \Gamma_0 \rangle$ & $-$ & $1.80$ & $3.39$ & $-$\\
$\langle \Gamma_+ / \Gamma_0 \rangle$ & $-$ & $1.08$ & $1.11$ & $-$\\
$N_+$ & $0$ & $5$ & $14$ & $0$\\
\hline \\[-2.8ex]

 & \multicolumn{4}{c}{\textit{2,000 words}} \\
\cline{2-5}
$\Gamma_0$ & $14.10\%$  & $19.23\%$  & $24.36\%$ & $30.77\%$ \\
$max\ \Gamma_+$ & $30.77\%$ & $25.64\%$ & $26.92\%$ & $34.62\%$ \\
$\langle \Gamma_+ - \Gamma_0 \rangle$ & $9.58$ & $1.92$ & $1.76$ & $3.08$\\
$\langle \Gamma_+ / \Gamma_0 \rangle$ & $1.68$ & $1.10$ & $1.07$ & $1.10$\\
$N_+$ & $19$ & $100$ & $8$ & $5$\\
\hline \\[-2.8ex]

& \multicolumn{4}{c}{\textit{2,500 words}} \\
\cline{2-5}
$\Gamma_0$ & $25.64\%$  & $32.05\%$  & $34.62\%$ & $41.03\%$ \\
$max\ \Gamma_+$ & $-$ & $43.59\%$ & $-$ & $48.72\%$ \\
$\langle \Gamma_+ - \Gamma_0 \rangle$ & $-$ & $5.33$ & $-$ & $4.27$\\
$\langle \Gamma_+ / \Gamma_0 \rangle$ & $-$ & $1.17$ & $-$ & $1.10$\\
$N_+$ & $0$ & $19$ & $0$ & $18$\\
\hline \\[-2.8ex]

& \multicolumn{4}{c}{\textit{5,000 words}} \\
\cline{2-5}
$\Gamma_0$ & $33.33\%$  & $46.15\%$  & $29.49\%$ & $62.82\%$ \\
$max\ \Gamma_+$ & $-$ & $-$ & $-$ & $65.38\%$ \\
$\langle \Gamma_+ - \Gamma_0 \rangle$ & $-$ & $-$ & $-$ & $1.71$\\
$\langle \Gamma_+ / \Gamma_0 \rangle$ & $-$ & $-$ & $-$ & $1.023$\\
$N_+$ & $0$ & $0$ & $0$ & $3$\\
\hline \\[-2.8ex]

& \multicolumn{4}{c}{\textit{10,000 words}} \\
\cline{2-5}
$\Gamma_0$ & $33.33\%$  & $67.95\%$  & $32.05\%$ & $88.46\%$ \\
$max\ \Gamma_+$ & $48.72\%$ & $73.08\%$ & $41.03\%$ & $-$ \\
$\langle \Gamma_+ \rangle$ & $8.02$ & $3.31$ & $4.70$ & $-$\\
$\langle \Gamma_+ / \Gamma_0 \rangle$ & $1.24$ & $1.05$ & $1.15$ & $-$\\
$N_+$ & $20$ & $17$ & $18$ & $0$\\
%\hline \\[-1.8ex]

\hline
\hline \\[-1.8ex]
\end{tabular}
\end{table}

%\singlespacing
\begin{table}[!htbp] \centering
   \caption{Statistics of performance obtained with \emph{word2vec}. The network construction phase \emph{considered} stopwords. $\Gamma_0$ is the the accuracy rate obtained with the traditional co-occurrence model and $max\ \Gamma_+$ is the highest accuracy rate considering different number of additional virtual edges. $\langle \Gamma_+ - \Gamma_0 \rangle$  and $\langle \Gamma_+ / \Gamma_0 \rangle$ are the average absolute and relative improvement in performance, respectively. $N_+$ is the total number of cases with an improvement in performance.}
  \label{tabdetfin}
  \footnotesize
\begin{tabular}{@{\extracolsep{5pt}}lcccc}
\\[-1.8ex]\hline
\hline \\[-2.8ex]
 & \multicolumn{4}{c}{\textit{1,000 words}} \\
\cline{2-5}
\\[-1.8ex] & DT & KNN & NB & SVM \\
%\\[-1.8ex] & (1) & (2)\\
\hline \\[-1.8ex]
$\Gamma_0$ & $16.67\%$ & $17.95\%$ & $17.95\%$ & $29.49\%$ \\
$max\ \Gamma_+$ & $25.64\%$ & $-$ & $24.36\%$ & $-$ \\
$\langle \Gamma_+ - \Gamma_0 \rangle$ & $4.43$ & $-$ & $3.65$ &$-$\\
$\langle \Gamma_+ / \Gamma_0 \rangle$ & $1.27$ & $-$ & $1.20$ & $-$\\
$N_+$ & $11$ & $0$ & $13$ & $0$\\
\hline \\[-2.8ex]
 & \multicolumn{4}{c}{\textit{1,500 words}} \\
\cline{2-5}
$\Gamma_0$ & $23.08\%$  & $21.79\%$  & $30.77\%$ & $37.18\%$ \\
$max\ \Gamma_+$ & $-$ & $26.92\%$ & $34.62\%$ & $-$ \\
$\langle \Gamma_+ - \Gamma_0 \rangle$ & $-$ & $2.85$ & $2.00$ & $-$\\
$\langle \Gamma_+ / \Gamma_0 \rangle$ & $-$ & $1.13$ & $1.07$ & $-$\\
$N_+$ & $0$ & $9$ & $16$ & $0$\\
\hline \\[-2.8ex]

 & \multicolumn{4}{c}{\textit{2,000 words}} \\
\cline{2-5}
$\Gamma_0$ & $14.10\%$  & $19.23\%$  & $24.36\%$ & $30.77\%$ \\
$max\ \Gamma_+$ & $30.77\%$ & $26.92\%$ & $30.77\%$ & $35.9\%$ \\
$\langle \Gamma_+ - \Gamma_0 \rangle$ & $10.26$ & $3.76$ & $3.93$ & $2.82$\\
$\langle \Gamma_+ / \Gamma_0 \rangle$ & $1.73$ & $1.20$ & $1.16$ & $1.09$\\
$N_+$ & $20$ & $15$ & $15$ & $15$\\
\hline \\[-2.8ex]

& \multicolumn{4}{c}{\textit{2,500 words}} \\
\cline{2-5}
$\Gamma_0$ & $25.64\%$  & $32.05\%$  & $34.62\%$ & $41.03\%$ \\
$max\ \Gamma_+$ & $26.92\%$ & $39.74\%$ & $42.31\%$ & $51.28\%$ \\
$\langle \Gamma_+ - \Gamma_0 \rangle$ & $1.28$ & $3.95$ & $4.20$ & $7.76$\\
$\langle \Gamma_+ / \Gamma_0 \rangle$ & $1.05$ & $1.23$ & $1.12$ & $1.19$\\
$N_+$ & $1$ & $13$ & $18$ & $20$\\
\hline \\[-2.8ex]

& \multicolumn{4}{c}{\textit{5,000 words}} \\
\cline{2-5}
$\Gamma_0$ & $33.33\%$  & $46.15\%$  & $29.49\%$ & $62.82\%$ \\
$max\ \Gamma_+$ & $-$ & $-$ & $-$ & $64.10\%$ \\
$\langle \Gamma_+ \rangle$ & $-$ & $-$ & $-$ & $1.28$\\
$\langle \Gamma_+ / \Gamma_0 \rangle$ & $-$ & $-$ & $-$ & $1.02$\\
$N_+$ & $0$ & $0$ & $0$ & $1$\\
\hline \\[-2.8ex]

& \multicolumn{4}{c}{\textit{10,000 words}} \\
\cline{2-5}
$\Gamma_0$ & $33.33\%$  & $67.95\%$  & $32.05\%$ & $88.46\%$ \\
$max\ \Gamma_+$ & $38.46\%$ & $75.64\%$ & $52.56\%$ & $-$ \\
$\langle \Gamma_+ - \Gamma_0 \rangle$ & $2.92$ & $3.93$ & $18.27$ & $-$\\
$\langle \Gamma_+ / \Gamma_0 \rangle$ & $1.09$ & $1.06$ & $1.57$ & $-$\\
$N_+$ & $11$ & $15$ & $20$ & $-$\\
%\hline \\[-1.8ex]

\hline
\hline \\[-1.8ex]
\end{tabular}
\end{table}

%\singlespacing
\begin{table}[!htbp] \centering
\caption{Statistics of performance obtained with \emph{FastText}. The network construction phase \emph{considered} stopwords. $\Gamma_0$ is the the accuracy rate obtained with the traditional co-occurrence model and $max\ \Gamma_+$ is the highest accuracy rate considering different number of additional virtual edges. $\langle \Gamma_+ - \Gamma_0 \rangle$  and $\langle \Gamma_+ / \Gamma_0 \rangle$ are the average absolute and relative improvement in performance, respectively. $N_+$ is the total number of cases with an improvement in performance..}
\label{tabdetfin}
\footnotesize
\begin{tabular}{@{\extracolsep{5pt}}lcccc}
\\[-1.8ex]\hline
\hline \\[-2.8ex]
 & \multicolumn{4}{c}{\textit{1,000 words}} \\
\cline{2-5}
\\[-1.8ex] & DT & KNN & NB & SVM \\
%\\[-1.8ex] & (1) & (2)\\
\hline \\[-1.8ex]
$\Gamma_0$ & $16.67\%$ & $17.95\%$ & $17.95\%$ & $29.49\%$ \\
$max\ \Gamma_+$ & $26.92\%$ & $-$ & $20.51\%$ & $-$ \\
$\langle \Gamma_+ - \Gamma_0 \rangle$ & $4.48$ & $-$ & $2.56$ &$-$\\
$\langle \Gamma_+ / \Gamma_0 \rangle$ & $1.27$ & $-$ & $1.14$ & $-$\\
$N_+$ & $18$ & $0$ & $3$ & $0$\\
\hline \\[-2.8ex]
 & \multicolumn{4}{c}{\textit{1,500 words}} \\
\cline{2-5}
$\Gamma_0$ & $23.08\%$  & $21.79\%$  & $30.77\%$ & $37.18\%$ \\
$max\ \Gamma_+$ & $-$ & $23.08\%$ & $33.33\%$ & $38.46\%$ \\
$\langle \Gamma_+ - \Gamma_0 \rangle$ & $-$ & $1.29$ & $1.71$ & $1.28$\\
$\langle \Gamma_+ / \Gamma_0 \rangle$ & $-$ & $1.06$ & $1.06$ & $1.03$\\
$N_+$ & $0$ & $5$ & $3$ & $4$\\
\hline \\[-2.8ex]

 & \multicolumn{4}{c}{\textit{2,000 words}} \\
\cline{2-5}
$\Gamma_0$ & $14.10\%$  & $19.23\%$  & $24.36\%$ & $30.77\%$ \\
$max\ \Gamma_+$ & $30.77\%$ & $21.79\%$ & $26.92\%$ & $35.90\%$ \\
$\langle \Gamma_+ - \Gamma_0 \rangle$ & $8.98$ & $2.05$ & $1.85$ & $3.53$\\
$\langle \Gamma_+ / \Gamma_0 \rangle$ & $1.64$ & $1.11$ & $1.08$ & $1.11$\\
$N_+$ & $20$ & $10$ & $9$ & $8$\\
\hline \\[-2.8ex]

& \multicolumn{4}{c}{\textit{2,500 words}} \\
\cline{2-5}
$\Gamma_0$ & $25.64\%$  & $32.05\%$  & $34.62\%$ & $41.03\%$ \\
$max\ \Gamma_+$ & $-$ & $35.90\%$ & $-$ & $48.72\%$ \\
$\langle \Gamma_+ - \Gamma_0 \rangle$ & $-$ & $2.33$ & $-$ & $4.49$\\
$\langle \Gamma_+ / \Gamma_0 \rangle$ & $-$ & $1.07$ & $-$ & $1.11$\\
$N_+$ & $0$ & $11$ & $0$ & $20$\\
\hline \\[-2.8ex]

& \multicolumn{4}{c}{\textit{5,000 words}} \\
\cline{2-5}
$\Gamma_0$ & $33.33\%$  & $46.15\%$  & $29.49\%$ & $62.82\%$ \\
$max\ \Gamma_+$ & $41.03\%$ & $47.44\%$ & $-$ & $65.38\%$ \\
$\langle \Gamma_+ - \Gamma_0 \rangle$ & $3.67$ & $1.29$ & $-$ & $1.92$\\
$\langle \Gamma_+ / \Gamma_0 \rangle$ & $1.11$ & $1.03$ & $-$ & $1.03$\\
$N_+$ & $7$ & $1$ & $0$ & $2$\\
\hline \\[-2.8ex]

& \multicolumn{4}{c}{\textit{10,000 words}} \\
\cline{2-5}
$\Gamma_0$ & $33.33\%$  & $67.95\%$  & $32.05\%$ & $88.46\%$ \\
$max\ \Gamma_+$ & $44.87\%$ & $73.08\%$ & $37.18\%$ & $-$ \\
$\langle \Gamma_+ - \Gamma_0 \rangle$ & $5.53$ & $2.70$ & $3.99$ & $-$\\
$\langle \Gamma_+ / \Gamma_0 \rangle$ & $1.17$ & $1.04$ & $1.12$ & $-$\\
$N_+$ & $16$ & $19$ & $18$ & $0$\\
%\hline \\[-1.8ex]

\hline
\hline \\[-1.8ex]
\end{tabular}
\end{table}

%$\Gamma_0$: taxa de acerto sem adicionar arestas;

%$\langle \Gamma_+ \rangle$: aumento médio do acerto (considerando apenas os casos que melhorou). \\

%$\langle \Gamma_+ / \Gamma_0 \rangle$: aumento relativo médio (considerando apenas os casos que melhorou)

%$N+$: quantos casos teve aumento em performance (dos casos de 1\% a 20\%)

\end{document}